\documentclass[runningheads]{llncs}
\usepackage{graphicx}
\usepackage{amssymb}
\usepackage{amsmath}
\usepackage{bibnames}
\usepackage{booktabs}
\usepackage{todonotes}
\usepackage{makecell}
\usepackage[caption=false]{subfig}
\renewcommand{\arraystretch}{1.2}

\newcommand{\ra}[1]{\renewcommand{\arraystretch}{#1}}
\newcommand{\vect}[1]{\mathbf{#1}}
\newcommand{\pSeq}{S}
\newcommand{\pMap}{P}

\newcommand{\pMapVoxel}{p_{x,y,z}}
\newcommand{\pMapPredVoxel}{\hat{p}_{x,y,z}}
\newcommand{\pSeqVoxelSeq}{\vect{s}_{1:T,x,y,z}}
\newcommand{\modelName}{\emph{A}}
\newcommand{\uncertaintyPred}{\hat{b}_{x,y,z}}
\newcommand{\varPred}{\hat{\sigma}^2}

\begin{document}
\title{Synthetic Perfusion Maps: Imaging Perfusion Deficits in DSC-MRI with Deep Learning.\thanks{Supported by SNSF grant no. 320030L\textunderscore170060 and the Swiss Heart Foundation.}}
\titlerunning{Synthetic Perfusion Maps with Deep Learning}
\author{Andreas Hess\inst{1} \and
Raphael Meier\inst{1} \and
Johannes Kaesmacher\inst{1,2} \and
Simon Jung\inst{2} \and
\\Fabien Scalzo\inst{3} \and
David Liebeskind\inst{3} \and
Roland Wiest\inst{1} \and
Richard McKinley\inst{1}}
\authorrunning{A Hess et al.}
\institute{Support Center for Advanced Neuroimaging, Institute for Diagnostic and Interventional Neuroradiology, Inselspital, University of Bern, Switzerland \and
Department of Neurology, Inselspital, University of Bern, Switzerland \and
Department of Neurology, University of California Los Angeles (UCLA), USA}
\maketitle              % typeset the header of the contribution
\begin{abstract}
In this work, we present a novel convolutional neural network based method for perfusion map generation in dynamic susceptibility contrast-enhanced perfusion imaging. The proposed architecture is trained end-to-end and solely relies on raw perfusion data for inference. We used a dataset of 151 acute ischemic stroke cases for evaluation. Our method generates perfusion maps that are comparable to the target maps used for clinical routine, while being model-free, fast, and less noisy.

% \keywords{Deep convolutional neural network \and Dynamic susceptibility contrast perfusion imaging  \and Ischemic stroke}
\end{abstract}
\section{Introduction}
\par{Dynamic susceptibility contrast-enhanced (DSC) magnetic resonance imaging (MRI) is an essential tool to assess perfusion deficits in acute ischemic stroke \cite{Olivot2009GeographyDEFUSE.}. Given a perfusion sequence, traditional methods rely on an estimation of the arterial input function (AIF) and on a parametric model to generate perfusion maps. Automatic estimation of the AIF tends to be non-robust, whereas a manual selection of the AIF is impractical. Besides, the computation of perfusion maps costs time in the range of minutes, in situations where time is often critical.}
\par{In 2017, Song et al. \cite{Song2017TemporalStroke} introduced temporal similarity perfusion (TSP) maps, together with a model-free, iterative process for their generation. The generated maps are compared to traditional time-to-peak (TTP) and mean-transit-time (MTT) maps to assess their clinical value for the detection of perfusion deficits. The proposed method operates without the need for AIFs, but it is fixed to the generation of TSP maps. Machine learning has been used extensively to post-process perfusion maps, e.g., to estimate tissue at risk (penumbra) \cite{McKinley2017FullyFASTER}. Meanwhile, in McKinley et. al~\cite{McKinley2018PerfusionSubmitted} regression of perfusion maps from raw DSC-MRI perfusion data in the presence of an externally provided AIF was demonstrated.  Fully automatic end-to-end regression of DSC perfusion maps has not been approached so far.}
\par{In this paper, we present a model-free, convolutional neural network (CNN) based architecture to predict arbitrary perfusion maps in an end-to-end manner. There have already been applications of CNNs on dynamic contrast-enhanced (DCE) perfusion sequences \cite{Ulas2018DirectLoss}. However, our work is the first application on DSC-MRI perfusion sequences to the best of our knowledge.}

\section{Methods}
\label{sec:methods}
\par{In this section, we describe the proposed neural network architecture for predicting synthetic perfusion maps. Since the architectures we developed for predicting different types of perfusion maps are similar (see supplemental material), we solely focus on the one used for predicting $T_{max}$ perfusion maps.}

\subsection{Pre-Processing and Augmentation}
\par{We pre-process both the raw perfusion data and the target perfusion maps. Each volume is padded with zeros to match the maximum size of any volume in the training dataset. Additionally, the perfusion sequence is complemented with volumes at the end to match the maximum sequence length of any perfusion sequence in the training dataset. For this, we reflect the data to generate frames for padding instead of using zero-filled volumes. The resulting perfusion maps are of size $24 \times 256 \times 256$, the perfusion sequences of size $80 \times 24 \times 256 \times 256$, where $80$ is the number of frames and $24$ the number of image slices.}
\par{After padding, we perform data standardization, i.e., we transform the data, so it has zero mean and unit variance. Finally, we apply voxel-wise temporal Gaussian smoothing for the perfusion sequence with $\sigma_{t} = 1.0$.}
\par{From inspecting the training portion of our dataset, we conclude that the time of perfusion sequence start is not in a global relation to the time of bolus arrival in the brain, i.e., there are cases where the bolus arrives much earlier or later than in most other cases. Even though such cases occur rarely, we want our model to be able to handle arbitrary delays of the bolus. To prevent the model from learning a global bolus arrival time, we augment the training dataset by randomly offsetting the perfusion sequence by -5 to 30 frames. A negative number means we remove frames at the beginning of the sequence and add padding frames at the end, a positive number indicates the opposite. The padding is generated via reflecting.}

\subsection{Deep Architecture for Perfusion Map Regression}
\label{subsec:methods:model}
\begin{figure}
  \centering
    \includegraphics[width=0.85\textwidth]{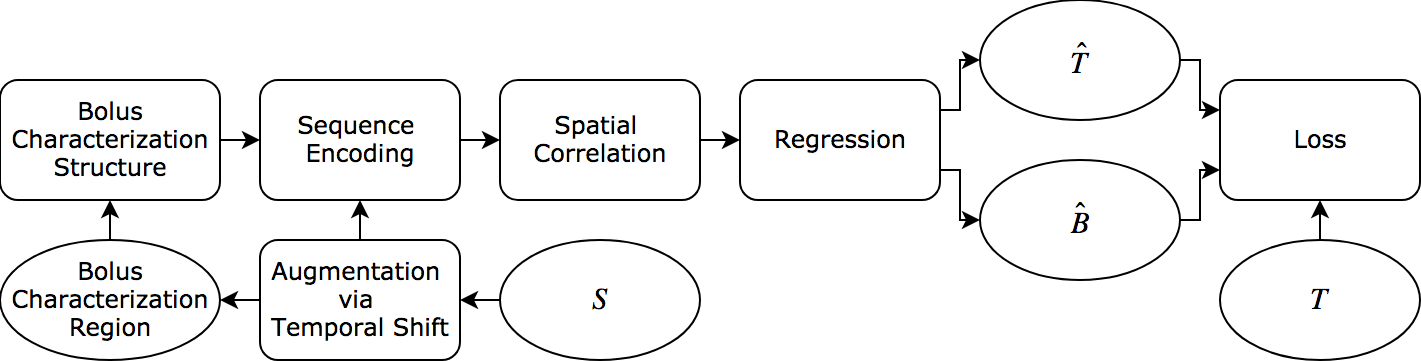}
  \caption{An overview of the perfusion map prediction model.}
  \label{fig:abstractArchitecture}
\end{figure}
\par{The goal is to predict voxel-values in a target perfusion map $\pMap$ based on the raw perfusion sequence $\pSeq$. Our crucial assumption is that a voxel-value $\pMapVoxel$ in perfusion map $\pMap$ mostly depends on the sequence of voxels at the same location in the raw perfusion sequence, i.e., on $\pSeqVoxelSeq$, where $T$ indicates the total number of frames in the perfusion sequence. There are obvious limitations to this assumption, which will be discussed in Section \ref{subsec:results:discussion}. Based on this assumption, we use a CNN to capture the temporal evolution of the raw perfusion sequence voxel-wise. Figure \ref{fig:abstractArchitecture} shows an overview of our architecture. The individual substructures will be briefly explained in the following sections.}

\subsubsection*{Bolus Characterization.}
\par{Our model has neither knowledge of the arterial input function nor the time of bolus arrival in the brain. Ignoring these aspects would significantly impair the performance of our model. Therefore, we add a bolus characterization structure (BCS) which should help capture the time of bolus arrival in the brain as well as the AIF. Guided by the fact that those characteristics are captured best by large blood vessels entering the brain, we select the input to the BCS to be a patch sequence from the perfusion sequence, located at the transition between the basilar artery and the posterior cerebral artery. The location of this patch is globally fixed, i.e., it is not fine-tuned to the individual volume. Therefore, it may happen that this patch does not contain the desired blood vessels for specific instances in our data. The BCS processes the supplied patch sequence via two 3D convolutional layers, encoding each patch into a vector of size 16. The sequence of encoded patches is forwarded to the sequence encoder.}

\subsubsection*{Sequence Encoding.}
\par{The sequence encoder handles every voxel sequence $\pSeqVoxelSeq$ independently, together with the additional information supplied by the bolus characterization structure. Note that this information is the same for all $\pSeqVoxelSeq$ of a perfusion sequence $\pSeq$. For simplicity, we will describe how the sequence encoder handles one individual voxel sequence. In reality, the sequence encoder processes multiple voxel sequences concurrently.}
\par{Effectively, the sequence encoder works on three inputs: a sequence of voxel-values $\pSeqVoxelSeq$, the sequence of frame times $\in \mathbb{R}^{80}$, and the sequence of encoded patches from the bolus characterization structure. These sequences are concatenated along their non-temporal dimension, resulting again in a sequence of length $80$. This result is passed through three 1D convolutional layers, of which the first two are followed by a max-pooling layer. The output is a vector of size 256, capturing the evolution over time of one voxel in the perfusion sequence.}
\begin{figure}
  \centering
    \includegraphics[width=0.6\textwidth]{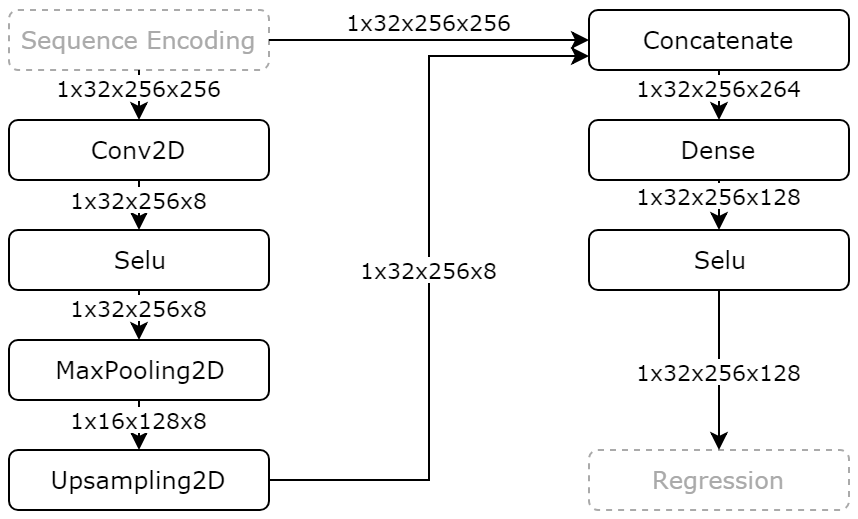}
  \caption{The architecture we used for local spatial correlation. The dense layer only operates on the last dimension of the data. Note that the shape of the data passed from the sequence encoder to the 2D convolution stems from the fact that we process the volumes in patches of $1 \times 32 \times 256$ and that each voxel sequence $\pSeqVoxelSeq$ is encoded into a vector of size 256. The first dimension of the data is redundant and is only depicted for clarity. \emph{Selu} denotes the activation function used \cite{Klambauer2017Self-NormalizingNetworks}.}
  \label{fig:spatialCorrelation}
\end{figure}
\subsubsection*{Spatial Correlation.}
\par{So far, there is no flow of information between neighboring voxels, i.e., each voxel sequence $\pSeqVoxelSeq$ is handled independently. Due to the low spatial resolution of the volumes along the axial axis, we omit spatial correlation between slices. To allow for some learned local filtering within slices, we apply 2D convolution to the voxel-wise output of the sequence encoder. Figure \ref{fig:spatialCorrelation} illustrates this process.}

\subsubsection*{Regression.}
\par{Given the spatially correlated encoding of voxel sequence $\pSeqVoxelSeq$, the prediction of the perfusion map voxel-value $\pMapPredVoxel$ and the estimated uncertainty $\uncertaintyPred$ is performed via a fully connected layer with two output neurons and identity activation.}

\subsubsection*{Loss.}
\par{The training objective is based on heteroscedastic aleatoric uncertainty modeling \cite{Kendall2017WhatVision}, i.e., instead of a single prediction, the model outputs a mean and a measure of uncertainty per voxel. This effectively corresponds to a probability distribution per voxel. The loss is given by the negative log-likelihood of the observed target map data. We choose a Laplace distribution to assign probabilities to observed values since a Gaussian distribution is too light tailed to cope with the amount of noise in target perfusion maps. Equation \ref{eq:uncertaintyLoss} formally defines the negative log-likelihood $l'$, given $\pMapVoxel$, $\pMapPredVoxel$ and estimated uncertainty parameter $\uncertaintyPred$. 
\begin{align}
l' \left( \pMapVoxel, \pMapPredVoxel, \uncertaintyPred \right) &= \log \uncertaintyPred + \frac{\left| \pMapVoxel - \pMapPredVoxel \right|}{\uncertaintyPred} \label{eq:uncertaintyLoss}
\end{align}
To be able to further focus the networks efforts onto voxel-values of high clinical importance and to simultaneously reduce the influence of noise, we apply a weighting scheme to the negative log-likelihood $l'$ of each voxel. The weighting scheme is based on a predefined function $I$ that assigns an importance to each value in $\pMap$ as shown in Equation \ref{eq:targetWeight}. The weight for a pair $\left( \pMapVoxel, \pMapPredVoxel \right)$ is given by the function $W$ as shown in Equation \ref{eq:lossWeight}. The final voxel-wise loss $l$ is given by Equation \ref{eq:loss}. We use an average to compute the loss over multiple voxels.}
\begin{align}
I(z) &= \left\{\begin{array}{ll}
        1.0, & \text{    if } 0.0 \leq z \leq 40.0\\
        0.1, & \text{    else}
        \end{array}\right. \label{eq:targetWeight} \\
W \left( \pMapVoxel, \pMapPredVoxel \right) &= \max_{z=\min(\pMapVoxel, \pMapPredVoxel)}^{\max(\pMapVoxel, \pMapPredVoxel)} I(z) \label{eq:lossWeight} \\
l \left( \pMapVoxel, \pMapPredVoxel, \uncertaintyPred \right) &= l' \left( \pMapVoxel, \pMapPredVoxel, \uncertaintyPred \right) \cdot W \left( \pMapVoxel, \pMapPredVoxel \right) \label{eq:loss}
\end{align}

\subsubsection*{Training.}
\par{The perfusion maps are processed in patches of size $1 \times 32 \times 256$, the perfusion sequences in patches of size $80 \times 1 \times 32 \times 256$. These patches are gathered in batches of size four before being passed to the network. For optimization, we use Adam \cite{Kingma2014Adam:Optimization} with an initial learning rate of $5e^{-4}$. The learning rate is divided by two every four epochs. We rely on dropout regularization with a dropout rate of $0.5$ for fully connected layers and do not use $l_2$-norm weight decay.}

\section{Experimental Setup and Results}
\par{The model was trained and evaluated on DSC-MRI perfusion data of patients with acute ischemic stroke. For a given perfusion sequence, the target perfusion maps were generated using \emph{oscillation index singular value decomposition} in Olea Sphere$^{\tiny{\textregistered}}$ 2.3 with default settings as used in clinical routine. The complete dataset contains 189 cases, of which we excluded 38 because the detected AIF was inaccurate, leaving us with a dataset of 151 cases. Approval for this retrospective study was obtained from the local ethics committee (KEK Bern, Switzerland, approval number: 231/14). Written informed consent was waived according to the retrospective nature of this analysis.}
\par{We randomly partitioned the dataset into training set, validation set and test set with ratios of $0.5$, $0.2$ and $0.3$ respectively. The training set was used for optimizing the network's weights, the validation set for model selection, and the test set for evaluating the final model.}
\par{The neural network was trained end-to-end, using the pre-processed perfusion sequences as input and the pre-processed perfusion maps from Olea as target maps. Training was performed on a Windows machine with an Intel Xeon E5-1630 v3 @ 3.7GHz and a Nvidia GeForce GTX 1080, and on an Ubuntu machine with an Intel i7-4790K CPU @ 4.00GHz and two Nvidia GeForce GTX 1070. All models were trained for 30 epochs, the selection of the final model was based on its performance on the validation set. The forward propagation took time in the range of seconds to compute a complete perfusion map.}

\subsection{Quantitative Results}
\begin{table*}\centering
\ra{1.1}
\begin{tabular}{@{}lll@{}}\toprule
\textbf{Model} & \textbf{MAEC@Validation} & \textbf{MAEC@Test}\\ \midrule
\textbf{Model \modelName{}} & \textbf{0.513} & 0.530 \\
\textbf{$-$ Augmentation} (\emph{B}) & 0.531 & \textbf{0.524} \\
\textbf{$-$ Spatial Correlation} (\emph{C}) & 0.562 & 0.629 \\
\textbf{$-$ Bolus Characterization} (\emph{D}) & 0.632 & 0.680 \\
\textbf{$-$ Loss Weighting} (\emph{E}) & 0.683 & 0.738 \\
\bottomrule
\vspace{2.0mm}
\end{tabular}
\caption{The MAEC of the predicted $T_{max}$ maps for different models. The lines below model \modelName{} show the performance of models where the listed component was removed.}
\label{tab:quantResults}
\end{table*}
\par{To quantitatively evaluate and compare different models, we used a mean absolute error with clipping (MAEC) as performance measure. It is identical to a mean absolute error, except that voxel-values are clipped to the interval $[0, 20]$ before computing differences. The clipping was done since values below $0$ mostly correspond to air and the ones above $20$ definitely indicate a perfusion deficit or are part of the noise. The exact values $0$ and $20$ were chosen because $[0, 20]$ is a reasonable window for inspecting $T_{max}$ maps.}
\begin{figure}
  \centering
    \includegraphics[trim={0 0.3cm 0 0.2cm},clip,width=0.5\textwidth]{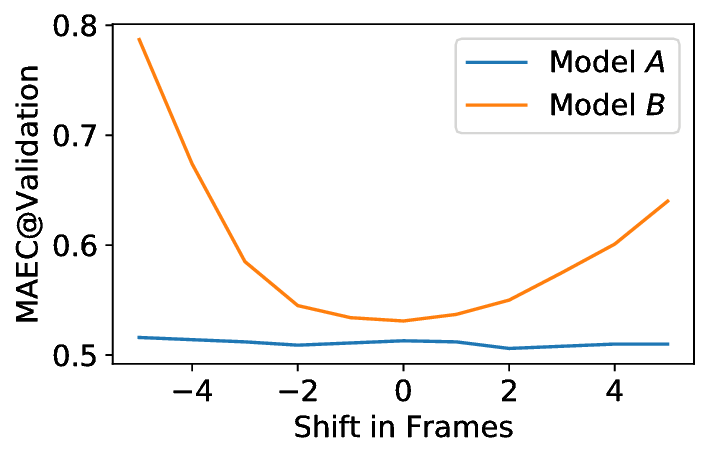}
  \caption{The influence on the MAEC for different models when shifting the raw perfusion sequence by a given number of frames.}
  \label{fig:timeShiftError}
\end{figure}
\par{The model described in Section \ref{subsec:methods:model} is referred to as model \modelName{} in the remainder of this section. Table \ref{tab:quantResults} shows the performance of model \modelName{} and compares it to variants of model \modelName{} where critical components introduced in Section \ref{subsec:methods:model} were removed. While model \modelName{} performed best on the validation set, model \emph{B} performed best on the test set. When manually inspecting the perfusion sequences of the validation and test set, we observed that the time of contrast bolus arrival varies more in the validation set than it does in the test set, with a standard deviation of 2.76 frames in contrast to 2.09 frames. Hence, the cases in the validation set did potentially benefit more from data augmentation via temporal shift. However, this temporal shift made training harder, which made the model perform slightly worse on cases where the bolus arrival delay was close to the mean delay. To further investigate the effectiveness of our augmentation, we measured the MAEC on the validation data for model \modelName{} and model \emph{B} after shifting the perfusion sequences by a number of frames. The results are shown in Figure \ref{fig:timeShiftError} and clearly indicate that the augmentation successfully helped the model to compensate for different times of bolus arrival.} 
\par{Further, we observed that removing the spatial correlation, the bolus characterization structure or the loss weighting significantly decreased the model's prediction accuracy on both the validation and the test set. It is evident that removing loss weighting increases the MAEC, since the loss weights assign high importance to the values that influence the MAEC.}

\subsection{Qualitative Results}
\par
\begin{figure}
  \centering
    \includegraphics[trim={0 0.6cm 0 0.1cm},clip,width=0.85\textwidth]{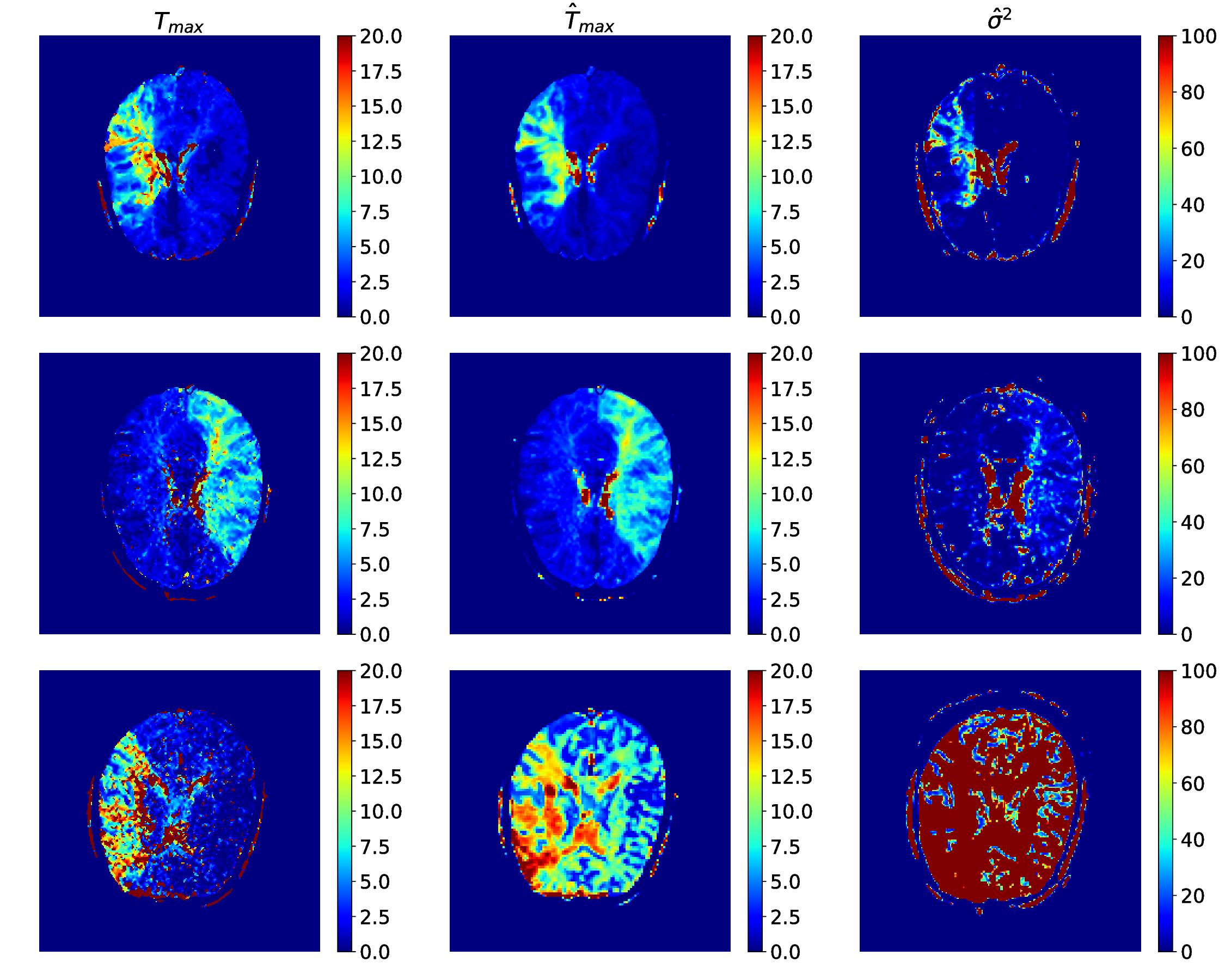}
  \caption{The target map $T_{max}$, the predicted map $\hat{T}_{max}$ and the estimated variance $\hat{\sigma}^2$ of the prediction for three examples from the test set. Note that $\hat{\sigma}^2$ is given by $2\hat{b}^2$.}
  \label{fig:targetPredictionVariance}
\end{figure}

{Figure \ref{fig:targetPredictionVariance} shows the target map $T_{max}$, the model's prediction $\hat{T}_{max}$ and the estimated variance of the prediction $\hat{\sigma}^2$ for three samples from the test set.}
\par{For the first two cases, the model's prediction is very close to the target map. Compared to the target maps, the predictions tend to contain less high-valued noise and generally have a smoother appearance. We would have expected the model to predict high uncertainty where its prediction error is likely to be high, i.e., we would have hoped for a positive correlation between prediction error and $\varPred$. The $\varPred$ maps do not match those expectations. Instead, we observed a correlation between the predicted value and $\varPred$, meaning the prediction of correct high values seems to be hard for the model. This observation makes sense since there is a considerable amount of high-valued noise in the target maps.}
\par{The third row of Figure \ref{fig:targetPredictionVariance} shows an example where our model failed. From observing the corresponding raw perfusion sequence and comparing it to perfusion sequences of examples where the model performed better, we noticed that the signal attenuation caused by the contrast agent is comparably weak for this case. Also, the signal is very noisy, partially due to slight head movements, which are amplified by the low axial resolution of the volumes. Given this additional information, we assumed that the bolus characterization structure was unable to correctly capture the bolus arrival in the brain, which led to a poor prediction.}

\section{Discussion and Conclusion}
\label{subsec:results:discussion}
\par{We made some simplifying assumptions in the presented approach, the most crucial one being that voxels in a perfusion map mainly depend on the perfusion sequence of voxels at the same location. This simplification does not always hold, especially when there was head movement during the sequence acquisition. An obvious solution to this is to register the individual volumes of the perfusion sequence before processing them any further. However, this is hard due to the low resolution of the volumes along the axial axis, which can lead to significant interpolation artifacts. Furthermore, it does not fit the concept of an end-to-end learning model. Another possible solution would be to make the sequence registration part of the model.}
\par{In conclusion, we presented a model-free, CNN-based method for inferring perfusion maps in an end-to-end manner. We demonstrated our method's performance on an ischemic stroke dataset of 151 patients and have shown that the predictions are comparable to the target perfusion maps. We are currently working on a clinical evaluation of the synthetic perfusion maps in order to confirm the applicability of CNNs in real-world DSC-MRI perfusion imaging.}

\bibliographystyle{splncs04}
\bibliography{Bibliography.bib}

\newpage

\section*{Appendix A: Model Hyperparameters}
\begin{table*}\centering
\ra{1.1}
\begin{tabular}{@{}ll@{}}\toprule
\textbf{Parameter Name} & \textbf{Value}\\ \midrule
\textbf{Input Dropout Rate} & 0.0 \\
\textbf{Convolutional Dropout Rate} & 0.5 \\
\textbf{Fully Connected Dropout Rate} & 0.0 \\
\textbf{$l_2$ Weight Decay $\lambda$} & 0.0 \\
\textbf{Learning Rate} & $5e^{-4}$ \\
\textbf{Optimizer} & Adam \\
\textbf{Sequence Encoding Dimension} & 256 \\
\textbf{Standardize Target Map} & yes \\
\textbf{Standardize Perfusion Sequence} & yes \\
\textbf{Temporal Augmentation Shift} & $\in [-5, 30]$ \\
\textbf{Kernel Initializer} & Xavier Uniform \\
\textbf{Activation Function} & Selu \\
\textbf{Batch Normalization} & no \\
\textbf{Error Model} & Negative Log-Likelihood via \\ & Heteroscedastic Aleatoric Uncertainty \\
\textbf{Temporal Smoothing} & Gaussian with $\sigma_t = 1.0$ \\
\textbf{Spatial Smoothing} & no \\
\bottomrule
\vspace{2.0mm}
\end{tabular}
\caption{A selection of hyperparameters in consideration, listed together with the respective value chosen for model \modelName{}.}
\label{tab:quantResults}
\end{table*}

\newpage

\section*{Appendix B: TTP and RBF}

\begin{figure}
    \centering
    \subfloat[\label{fig:ttpRbf:examples}]{\includegraphics[trim={0 0.6cm 0 0.1cm},clip,width=1.0\columnwidth]{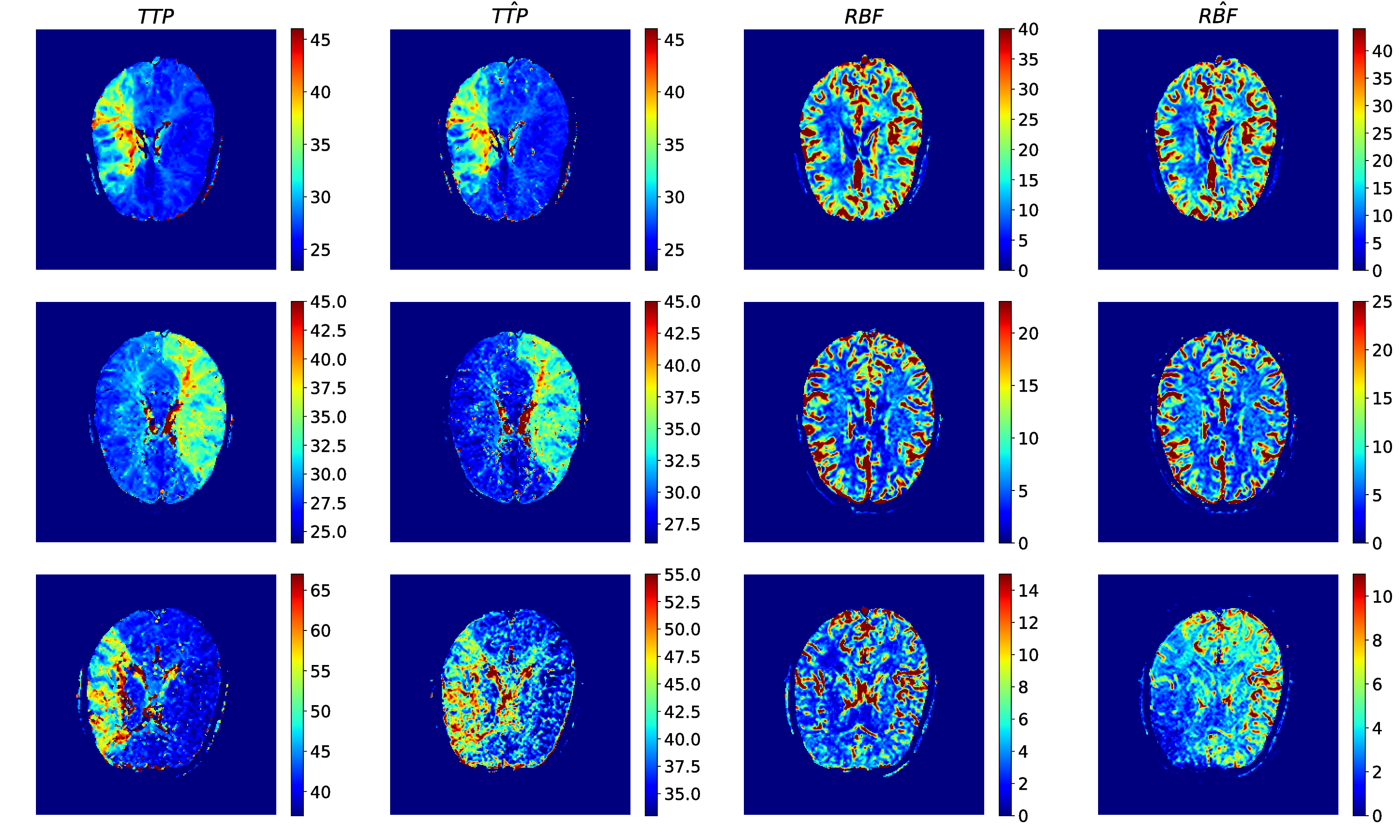}}
    \qquad
    \subfloat[\label{fig:ttpRbf:architecture}]{\includegraphics[width=1.0\columnwidth]{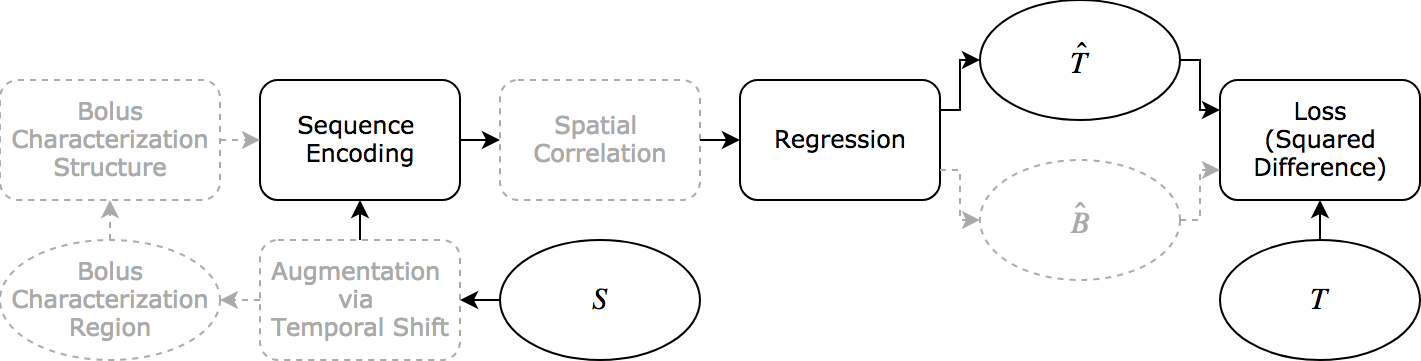}}
    \label{fig:ttpRbf}
	\caption{(a) The target maps $TTP$ and $RBF$ for three examples in the test set, together with the prediction of a network trained on the respective target maps. (b) An abstract outline of the architecture we trained to predict $TTP$ and $RBF$ maps. Compared to the architecture described in Section \ref{sec:methods}, the parts in gray were removed. Also, note that we use a squared loss instead of Laplacian-based negative log-likelihood. We trained two separate, identical estimators for the two target maps $TTP$ and $RBF$.}
\end{figure}

\end{document}